\documentclass{article}
\usepackage{spconf,amsmath,graphicx}
\usepackage{amsfonts}
\usepackage[table,xcdraw]{xcolor}
\usepackage[]{algorithm2e}
\usepackage{multirow}
\usepackage{placeins}
\usepackage{comment}
\usepackage{hyperref}
\usepackage{cite}
\usepackage{pgfplots}
\usepackage{siunitx}
\pgfplotsset{compat=1.9}
\usepgfplotslibrary{statistics}
\usetikzlibrary{pgfplots.groupplots}
\usetikzlibrary{patterns,positioning,fadings,through}
\usepackage{amsfonts}
\usepackage{multirow}
\usepackage{siunitx}
\usepackage[normalem]{ulem}
\usepackage{adjustbox}
\usepackage{balance}
\usepackage{makecell}
\usepackage{array}
\newcolumntype{?}{!{\vrule width 1.8pt}}
\newcommand{\PreserveBackslash}[1]{\let\temp=\\#1\let\\=\temp}
\newcolumntype{C}[1]{>{\PreserveBackslash\centering}p{#1}}
\newcolumntype{R}[1]{>{\PreserveBackslash\raggedleft}p{#1}}
\newcolumntype{L}[1]{>{\PreserveBackslash\raggedright}p{#1}}

\title{Training privacy-preserving video analytics pipelines by
suppressing features that reveal information about private attributes}

\name{Chau Yi Li, Andrea Cavallaro 
}
\address{Centre for Intelligent Sensing, Queen Mary University of London, UK
}
\begin{document}
%
\maketitle
\begin{abstract}
Deep neural networks are increasingly deployed for scene analytics, including to evaluate the attention and reaction of people exposed to out-of-home advertisements. However, the features extracted by a deep neural network that was trained to predict a specific, consensual attribute (e.g.~emotion) may also encode and thus reveal information about private, protected attributes (e.g.~age or gender). In this work, we focus on such leakage of private information at inference time. We consider an adversary with access to the features extracted by the layers of a deployed  neural network and use these features to predict private attributes. To prevent the success of such an attack, we modify the training of the network using a confusion loss that encourages the extraction of features that make it difficult for the adversary to accurately predict private attributes. We validate this training approach on image-based tasks using a publicly available dataset. Results show that, compared to the original network, the proposed PrivateNet can reduce the leakage of private information of a state-of-the-art emotion recognition classifier by 2.88\% for gender and by 13.06\% for age group, with a minimal effect on task accuracy. 
\end{abstract}
\begin{keywords}
feature learning, video analytics, privacy, private attributes. 
\end{keywords}

\section{Introduction}
\label{sec:intro}

Deep neural network classifiers are used to understand the reactions of audiences to advertisements in public spaces and during interactions with wayfinding kiosks. Such reactions are typically measured using computer vision by first detecting faces and then analysing  attributes associated with facial expressions. However, neural networks trained to extract features that predict a task-specific, consensual attribute (the target task), may also encode information about private attributes~\cite{Song2017ACM}, such as gender and age~\cite{PrivacyFromDemographics}. An adversary with access to the information extracted by the layers of a deep neural network can therefore use the pipeline for purposes different from the consensual ones. 


An adversary may perpetrate a {\em membership} or {\em attribute} inference attack. Privacy-preserving methods to protect from membership inference attacks focus on preventing the identification of individuals whose data were included in a training dataset~\cite{shokri2017membership, MembershipInference_2018, salem2018mlleaks_membership,melis2018exploiting_unintended_feature} or are ingested by a deployed system. Methods that protect from attribute inference attacks aim to prevent the prediction of (private) attributes from a training dataset~\cite{melis2018exploiting_unintended_feature} or from a classification pipeline deployed on a system, such as a smart kiosk. The latter scenario is the scope of our work: we consider a system whose untrusted rich execution environment is compromised by malware, but is equipped with security measures that ensure confidentiality and integrity of the visual data (e.g. a  limited-memory, isolated Trusted Execution Environment, or TEE). The memory limitation of a TEE can only host a few of the last layers of a feed-forward network, typically the layers after the backbone structure or only the fully-connected layer~\cite{Mo2020DarkneTZTM}. We consider the adversary to have access to the untrusted execution environment of the system and thus in a position to exploit the features extracted by these intermediate layers to infer private information~\cite{malekzadeh2021honestbutcurious}. Existing works that aim to prevent attribute inference attacks  modify the input data with adversarial perturbations~\cite{li2019scene} or train the network to cause mis-classification of private attributes, for instance using the cross-entropy loss~\cite{CensoringRepresentation_2016_Edwards_Storkey}.

In this work, we aim to prevent the estimation of private information at inference time by training the network to extract features that are  useful for the target task {only} and that cause a {known} adversary classifier to perform similarly to a {\em random classifier} on a protected attribute. To conceal the private information from the adversary classifier, we train the pipeline with a {confusion loss}. The proposed approach does not modify the network architecture or the number of its parameters, and hence has no latency impact on the inference pipeline.  In the specific implementation reported in this paper, we consider visual emotion recognition as the target task, and age and gender as protected attributes\footnote{The source code is available at the following URL:~\href{https://github.com/smartcameras/PrivateNet}{\url{https://github.com/smartcameras/PrivateNet}}.}. We also investigate the robustness of the proposed approach by training an adversary to classify the private information from the privacy-preserving network. 

\section{Method}

Let~$\mathcal{C}(\cdot)$ be a~$D$-class deep neural network classifier of $N$ layers that, given an image~$x$, predicts its {\em consensual} class as one in the set:
\begin{equation}
    \mathcal{Y} = \{y_1, y_2, \ldots, y_D\}.
\end{equation}
Let~$\mathcal{F}_i(x)$ be the output of layer~$i$ of~$C(\cdot)$, with~\mbox{$i\in\{1,...,N\}$}.  The output of the last layer~$N$ is a vector that represents the confidence of~$\mathcal{C}$ that~$x$ belongs to any  of the classes:
\begin{equation}
    \mathcal{F}_{\!N}(x) = (p^{y}_{1}, p^{y}_{2}, \ldots , p^{y}_{D}). 
    \label{eq:: consensual_confidence}
\end{equation} 

Let~$\mathcal{A}(\cdot)$ be an adversary~$K$-class classifier that aims to predict a  {\em protected, private} class from the output,~$\mathcal{F}_i(x)$,  of an intermediate layer as one in the set:
\begin{equation}
    \mathcal{S} = \{s_1, s_2, \ldots, s_K\}. 
\end{equation}

Let~$M$ be the number of layers in~$\mathcal{A}(\cdot)$. The output of the last layer~$M$  is a vector representing the confidence of~$\mathcal{A}(\cdot)$ that~$\mathcal{F}_i(x)$ belongs to any of the (private) classes:
\begin{equation}
    \mathcal{Q}_M(\mathcal{F}_i(x)) = (p^{s}_1, p^{s}_{2}, \ldots , p^{s}_{K}).
\end{equation}

Let~$\mathcal{X}$ be a set of face images, each annotated with an emotion (target) attribute,~$\hat{y}$, and a private attribute,~$\hat{s}$. We measure the leakage of private information as the accuracy,~$T_{\hat{s}}$, of the private attribute inference from~$\mathcal{X}$:
\begin{equation}
T_{\hat{s}} = 
    \frac{|\{x \in \mathcal{X}: \mathcal{A}(\mathcal{F}_i(x)) = \hat{s}\}|}{|\mathcal{X}|},
\end{equation}
where~$|\cdot|$ is the cardinality of a set. The higher $T_{\hat{s}}$, the higher the leakage of private information through~$\mathcal{C}(\cdot)$.

A classifier~$\mathcal{C}(\cdot)$ that is privacy-preserving should maintain a high accuracy in the prediction of the consensual task, while concealing the private attributes from the adversary classifier~$\mathcal{A}(\cdot)$, i.e.~$T_{\hat{s}}$ should be close to a random classifier.   

To predict the consensual class, the classifier~$\mathcal{C}(\cdot)$ is typically trained with a cross-entropy loss:
\begin{equation}
    \mathcal{L}_{CE}\Big(\hat{y}, \mathcal{F}_{\!N}(x)\Big) =  -\log \Bigg(\frac{exp(p_{\hat{y}})}{\sum_{i} exp(p^{y}_{i})}\Bigg),
    \label{eq: CE_loss}
\end{equation}
where~$exp(\cdot)$ is the exponential function.

To prevent the leakage of private information associated with a protected attribute,  we  propose to obfuscate with a confusion loss,~$\mathcal{L}_{con}$, the features in the intermediate layers that are useful to the adversary:
\begin{equation}
    \mathcal{L}_{con}\Big(\mathcal{F}_{\!i}(x)\Big) = \| \mathcal{Q}_{\!M}(\mathcal{F}_{\!i}(x))- \boldsymbol{\mathcal{U}}_D\|^2, 
     \label{eq: proposed_confusion_loss}
\end{equation}
where~$\boldsymbol{\mathcal{U}}_D = (\frac{1}{D},...,\frac{1}{D})$ denotes equal probability for each private attribute class, hence causing the adversary to perform similarly to a random classifier. As the features extracted by an intermediate layer embeds the features extracted by the layers preceding it, backpropagating the confusion loss through the entire network optimizes the layers preceding the targeted layer to extract generic features that cause the adversary to perform similarly to a random classifier on the protected attribute. Thus, in training we encourage the deep neural network to extract privacy-preserving features by combining Eq.~\ref{eq: CE_loss} and~\ref{eq: proposed_confusion_loss} as the overall loss function,~$\mathcal{L}$:
\begin{equation}
  \mathcal{L} = (1-\lambda)\mathcal{L}_{CE} + \lambda\mathcal{L}_{con},
  \label{eq: proposed_total_loss}
\end{equation}
where~$\lambda$ determines the relative importance between the losses:~\mbox{$\lambda  < 0.5$} gives more importance to maintaining the utility and~\mbox{$\lambda > 0.5$} gives more importance to confusing the adversary.

\begin{figure}[t!]
\centering
\begin{tabular}{cccc}
\includegraphics[width=0.95\columnwidth]{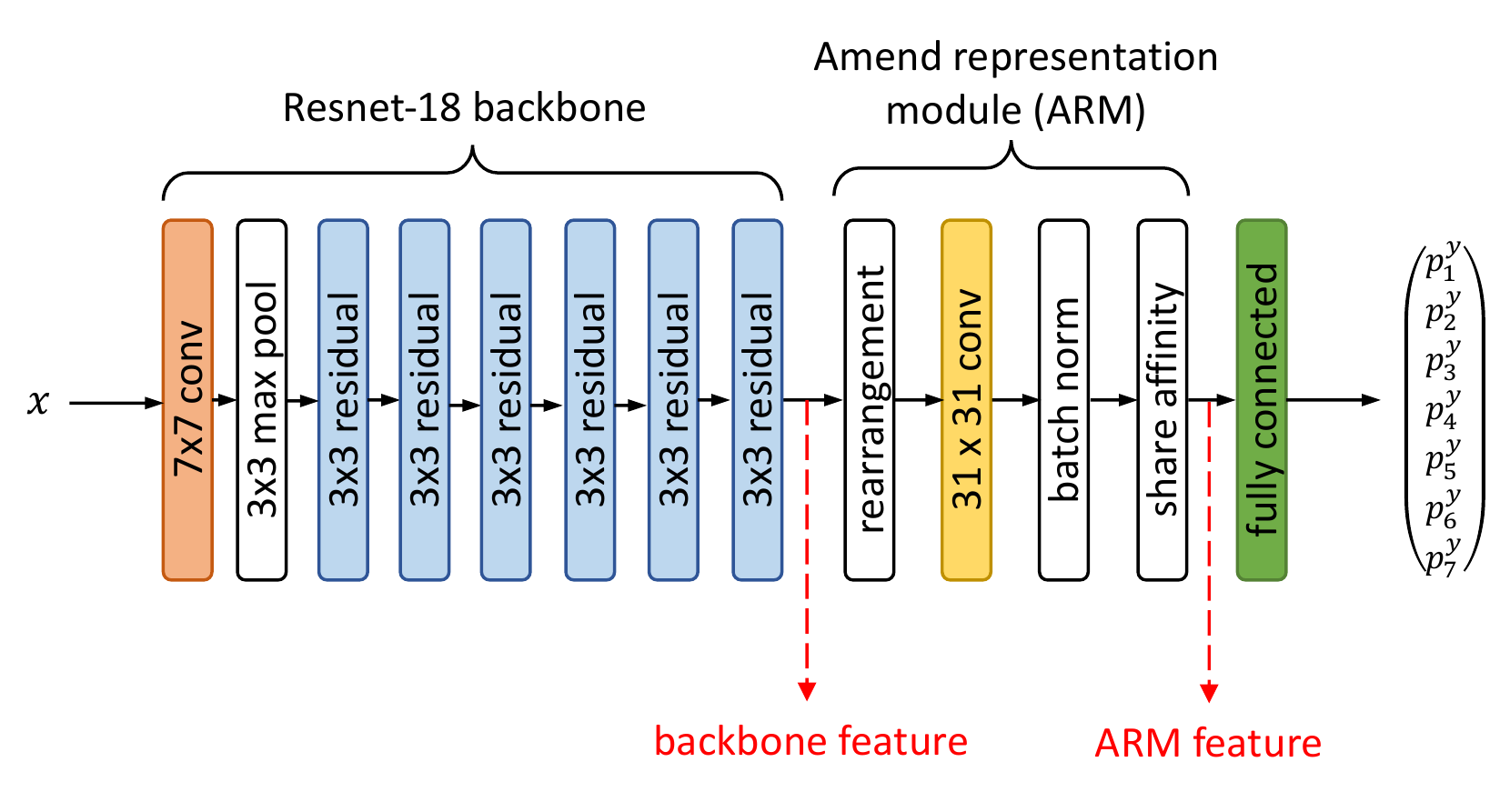}\\
\end{tabular}
\caption{Architecture of the ARM network~\cite{ARM_Shi_2021} used as the emotion recognition classifier. Given an image~$x$, the network outputs the probability~$p^y_i$, $ i\in\{1,..,7\}$, of each emotion class (Eq.~\ref{eq:: consensual_confidence}). An adversary could infer private, protected attributes from the features extracted by the intermediate layers of the network (red dashed arrows). Coloured blocks represent layers with trainable parameters.
 }
\label{fig:architecture}
\end{figure}  

We use as  dataset the Real-world Affective Faces Database (RAF-DB)~\cite{li2019reliable},  which contains 15,338 images. Each image was annotated, on average, by 40 annotators
into~\mbox{$D=7$} emotion attributes (surprise, fear, disgust, happiness, sadness, anger, and neutral) and demographics including race, gender and age group. Specifically, we focus on gender (\mbox{$K=2$}, male and female)\footnote{We excluded the images with {\em unsure} gender class, which contribute to~$6.3\%$ of the dataset. } and age group (\mbox{$K=5$}, namely 0-3, 4-19, 20-39, 40-69, and over 70). As for the emotion classifier~$\mathcal{C}(\cdot)$,  we use the Amend Representation Module network proposed by Shi and Zhou~\cite{ARM_Shi_2021}. On RAF-DB, the ARM network attains a state-of-the-art emotion recognition accuracy of 91.10\%. The ARM network aims to remove the distortion from the edges of the image on the features,  referred to as albino erosion. The ARM network consists of a ResNet-18~\cite{he2016deep} backbone, an Amend Representation Module (ARM), and a fully connected layer. The output of the ResNet-18 backbone is a feature map of size 7~$\times$~7 and 512 channels.
An ARM consists of 3 blocks, namely a feature rearrangement block, a convolution layer that is followed by batch normalisation, and a sharing affinity block. The rearrangement block distributes the backbone feature map into 2 channels, each of size 112~$\times$~112, while maintaining the relative positions between the features in the same channel~\cite{pixelshuffle}, whereas the sharing affinity block ensures the global average representation of faces in the mini-batch is propagated. 
For this paper, we choose to study the leakage of private information from the features extracted by the ResNet-18 backbone and that by the ARM, which represent layers that cannot be executed in the TEE.~Fig.~\ref{fig:architecture} shows the network  architecture and the features studied in this paper.

\section{Results}
 
In this section, we discuss the utility and the privacy level of the proposed privacy-preserving network, PrivateNet.
The {\em utility} of the  classifier network is its accuracy on emotion recognition. The {\em privacy level} provided for a private attribute by the network is how close the accuracy of an adversary classifier is to a random guess on that attribute.
Furthermore, the {\em robustness} of the network is the difference in accuracy between  the adversary accessing the features of the original network (Tab.~\ref{table: network_leakage_accuracy}) and the privacy-preserving networks. The more negative the percentage difference, the more robust the privacy-preserving network.

To establish a fair comparison with the private information that can be inferred from the original image and to ensure that any difference is due to the input and not to the change in the architecture, we use the same ARM network architecture in our experiments. The adversary for the backbone features consists of an ARM and a fully connected layer, whereas that for ARM features consists of a fully connected layer only. All networks are trained on an Nvidia Tesla V100-SXM2 GPU, using a learning rate of 0.001 with ADAM optimiser, with 200 epochs. The networks were tested on the same GPU.  


\begin{table}[t!]
\centering
\small
\setlength\tabcolsep{3pt}
\caption{The leakage of private information is measured as the accuracy of an adversary classifier that infers private attributes from the features extracted from the ARM network~\cite{ARM_Shi_2021}, trained for emotion recognition. Note that the accuracy of the adversary is lower than that of a network classifying the original image but higher than that of a random classifier.}
\vspace{5pt}
\begin{tabular}{lccccccl|l|}
\hline
\multirow{2}{*}{\textbf{Attribute}} &  {\textbf{Random}}  & \textbf{Baseline on}  &  \multicolumn{2}{c}{\textbf{Adversary on feature}}\\
\cline{4-5}
 &  \textbf{classifier} & \textbf{original image} & Backbone  & ARM  \\ \hline
Gender  & 50 & 88.22 &   74.97 &  63.30\\
Age  &  20 & 77.35 &  68.20 &  57.20\\
\hline
\end{tabular}
\label{table: network_leakage_accuracy}
\end{table}

Tab.~\ref{table: network_leakage_accuracy} reports the accuracy of the adversaries  on features extracted through different layers of the networks to infer two private attributes, namely gender and age.  A random classifier would achieve an accuracy of 50\%  for gender ($K=2$) and 20\%  for age ($K=5$). The classification accuracy of the ARM network architecture, using the original images as baselines, is 88.22\% for gender and 73.53\% for age.  The leakage of private information  decreases along the network, as the network extracts features that are more relevant to the target task. In particular, the accuracy of the adversaries on  the  gender attribute drops from 74.97\%, with features extracted by the backbone, to 63.30\%,  with features extracted by the ARM. Nonetheless, the adversary classifier can still predict the private attributes with high accuracy from the extracted features. For example, the adversary classifier who has access to the backbone feature can predict the gender attribute with 73.75\% accuracy (83.59\% of the baseline accuracy) and the age attribute with 68.20\% accuracy (88.17\% of the baseline).

Tab.~\ref{table: network_confusion_accuracy} compares the utility, privacy and robustness of 8 networks, trained with 2 loss functions, namely the proposed confusion loss and the adversarial loss by Edwards and Storkey~\cite{CensoringRepresentation_2016_Edwards_Storkey}, against 2 known adversaries on the backbone and ARM features, for 2 private attributes, namely gender and age. In the rest of the paper, we refer to the networks in the format~\textit{loss-feature-attribute}, hence~\textit{confusion-feature-attribute} is an example of the proposed PrivateNet. The  utility of the 8 networks ranges from 88.43\% to 89.86\%, corresponding to a slight drop (1.36\% to 2.93\%) from the original ARM network (accuracy of 91.10\%). Therefore training the network in a privacy-preserving way only slightly reduces the network performance for the consensual task.


\begin{table}[t!]
\small
\setlength\tabcolsep{2pt}
\caption{Utility, privacy and robustness of privacy-preserving networks trained with an adversarial loss~\cite{CensoringRepresentation_2016_Edwards_Storkey} and with the proposed confusion loss~(Eq.~\ref{eq: proposed_confusion_loss}). The level of privacy guaranteed for an attribute  is measured as the accuracy of an adversary classifier on the features of the network. The lower the accuracy, the more difficult to infer  private information from the features. Robustness is the difference in the accuracy of the adversary classifier on the features of the original network (Tab.~\ref{table: network_leakage_accuracy}) and the privacy-preserving network, reported in relative percentage difference. The more  negative the difference, the less susceptible (i.e.~more robust) the privacy-preserving network  to the adversary classifier. The results of the most robust network are shown in bold. KEY -- Att.: attribute; Adv.: adversarial; Prop.: proposed; K: known adversary; U: unknown adversary.}
\vspace{5pt}
\begin{tabular}{c}
     \begin{tabular}{cllccccccccccccc}
\hline
\textbf{Att.} & {\textbf{Feature}} & {\textbf{Loss}} & {\textbf{Utility}}  & \multicolumn{2}{c}{\textbf{Privacy}}    &  \multicolumn{2}{c}{\textbf{Robustness}} \\
\cline{5-8} 
& &  &  & K   & U & K   & U\\
\hline
\multirow{4}{*}{\rotatebox{90}{Gender}}& \multirow{2}{*}{Backbone} & Adv.  & 88.43 &   56.47 &  88.50 & -24.68\% & +18.04\% \\
& & Prop. & 89.40 & 51.27 & 72.81  & \textbf{-31.61\%} & \textbf{-2.88\%} \\ 
\cline{2-8}
& \multirow{2}{*}{ARM} & 
Adv. & 89.51 & 43.53 & 87.84 & \textbf{-36.17\%} & +37.99\%  \\
& &  Prop.  &  89.47 &   49.91 & 62.43 & -26.82\% & \textbf{+1.75\%}\\
\hline
\multirow{4}{*}{\rotatebox{90}{Age}} &  \multirow{2}{*}{Backbone} & Adv.  &  89.63 & 10.72 & 62.58 & \textbf{-83.06\%} & -8.24\% \\
&  & Prop. &   89.83 &  20.18 & 59.29 & -68.12\% & \textbf{-13.06\%}\\
\cline{2-8}
&  \multirow{2}{*}{ARM} & Adv. &  89.47 & 10.72 & 54.17 & \textbf{-81.26\%} & \textbf{-5.30\%}\\
& &  Prop.  & 89.86 & 21.28 & 56.88 & -62.80\% & -0.56\% \\
\hline
\end{tabular}
\end{tabular}
\label{table: network_confusion_accuracy}
\end{table}

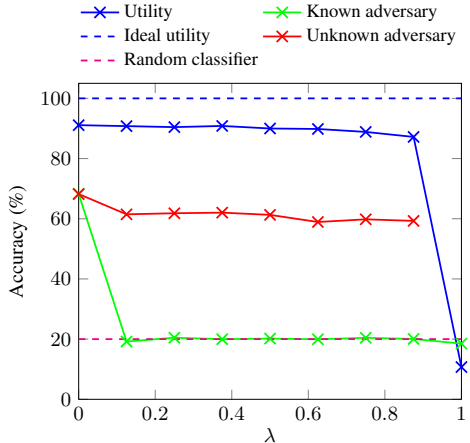
\begin{figure}[t!]
\centering
\begin{tikzpicture}[scale=0.85, every node/.style={scale=0.85}]
\pgfplotsset{compat=1.3}
 \tikzstyle{every node}=[font=\small, outer sep=0,trim axis right]
   \begin{axis}[
    xlabel={$\lambda$},
    ylabel={Accuracy (\%)},
    ylabel shift={-1pt},
    xlabel shift={-3pt},
    ytick={0,20,40,60,80,100},
    xmin=0,  xmax=1, ymax=105, ymin=0, legend cell align={left}, 
    legend columns=2,
   legend style={at={(0.5,1.02)},anchor=south, nodes={scale=0.92}, {draw=none}},
 width=0.88\columnwidth
  ]
\addplot+[color=blue,mark=x, mark size=3.5,thick]table[x=x,y=Utility] {
x		Utility		Known		Unknown
0		91.1		68.22			
0.125		90.78		19.2		61.47	
0.25		90.45		20.47		61.83	
0.375		90.84		19.98		62.03	
0.5		89.99		20.18		61.29	
0.625		89.83		19.95		58.92	
0.75		88.85		20.44		59.81	
0.875		87.19		20.05		59.29	
1		10.72		18.51			
}; \label{plots:utility}
\addlegendentry{Utility}
\addplot+[color=green,mark=x, mark size=3.5,thick]table[x=x,y=Known] {
x		Utility		Known		Unknown
0		91.1		68.22			
0.125		90.78		19.2		61.47	
0.25		90.45		20.47		61.83	
0.375		90.84		19.98		62.03	
0.5		89.99		20.18		61.29	
0.625		89.83		19.95		58.92	
0.75		88.85		20.44		59.81	
0.875		87.19		20.05		59.29	
1		10.72		18.51			
}; \label{plots:known_adversary}
\addlegendentry{Known adversary}
\addplot[blue,dashed,thick] 
coordinates {(0,100) (1,100)}; 
\label{plots:ideal_utility}
\addlegendentry{Ideal utility}
\addplot+[color=red,mark=x, mark size=3.5, thick]table[x=x,y=Unknown] {
x		Utility		Known		Unknown
0		91.1		68.22		68.22
0.125		90.78		19.2		61.47	
0.25		90.45		20.47		61.83	
0.375		90.84		19.98		62.03	
0.5		89.99		20.18		61.29	
0.625		89.83		19.95		58.92	
0.75		88.85		20.44		59.81	
0.875		87.19		20.05		59.29	
}; \label{plots:unknown_adversary}
\addlegendentry{Unknown adversary}
\addplot[magenta,dashed,thick] 
	coordinates {(0,20) (1,20)}; \label{plots:random_guess}
	\addlegendentry{Random classifier}
  \end{axis}
\end{tikzpicture}
\caption{Impact of the relative weighting of the cross-entropy loss and the proposed confusion loss ($\lambda$~in~Eq.~\ref{eq: proposed_total_loss}) on the accuracy of the consensual task (utility) and of adversaries that use the backbone features to predict the age group ($K=5$). Note that~$\lambda=1$ means that the network is not trained for the target task and therefore we do not report the accuracy of the unknown adversary for this value. Dashed lines show the ideal behaviours, i.e.~maximum utility for the consensual attribute and random results for the private attribute. 
}
\label{fig: lambda_test}
\end{figure} 

For the binary (\mbox{$K=2$}) gender attribute, the accuracy of the known adversary on PrivateNet is 51.27\% on~\textit{confusion-backbone-gender} and 49.91\% on~\textit{confusion-ARM-gender}. Thus the confusion loss has achieved the goal of obfuscating the known adversary to perform similarly to a random classifier (50\%). However, 
the high accuracy of the unknown adversary (72.81\% and 62.43\% on backbone and ARM features, respectively) indicates that the networks are not robust against the attack of an adversary that is unknown at the time of training. Training the network with adversarial loss forces the known adversary to mis-classify the binary attribute (from male to female and from female to male), hence the accuracy of the known adversary on \textit{adversarial-backbone-gender} is lower than that on \textit{confusion-backbone-gender} (43.53\%). However, the  accuracy of an unknown adversary on \textit{adversarial-backbone-gender} (87.84\%) is higher than that of \textit{confusion-backbone-gender}, and also higher than that of the original network (68.20\%). Moreover, while training with the confusion loss has increased the robustness by 2.88\% on \textit{confusion-backbone-gender} and decreased it by 1.75\% on \textit{confusion-ARM-gender}, training with the adversarial loss decreases the robustness against an unknown adversary by 18.04\% and 37.99\% on \textit{adversarial-backbone-gender} and \textit{adversarial-ARM-gender}, respectively. Therefore, training with an adversarial loss in fact encourages the network to extract features that are more useful for inferring the binary attribute, hence reducing the ability of the network to protect the attribute. 

For the age group attribute (\mbox{$K=5$}), training with the adversarial loss  protects the private attribute against a known adversary better than training it with the confusion loss: the accuracy of the adversary is 10.72\% on  \textit{adversarial-backbone-age} and 20.18\% on \textit{confusion-backbone-age}. However, the accuracy of an unknown adversary is 62.58\% on \textit{adversarial-backbone-age} and 59.29\% on \textit{confusion-backbone-age}. Overall, the proposed PrivateNet are  more  robust than networks trained with the adversarial loss.

Fig.~\ref{fig: lambda_test} reports the results obtained when  varying~the relative weight of the cross-entropy  and confusion losses ($\lambda$ in Eq.~8) on the consensual task and the robustness of the proposed privacy-preserving network. Note that~$\lambda=0$ is equivalent to the original ARM network that is optimised for emotion recognition only;  increasing~$\lambda$ gives more importance to the confusion loss; and~$\lambda=1$ discards the cross-entropy loss and hence the network is optimised for misleading the known adversary only. As~$\lambda$ increases, the utility (emotion recognition accuracy) decreases, as expected. While training  with most values of~$\lambda$ can cause the accuracy of the known adversary  to be close to that of a random classifier, the robustness of the PrivateNet increases with~$\lambda$. Tab.~\ref{table: inference_time} reports the  inference time of the proposed networks on RAF-DB. As the numbers of parameters are the same, the inference times of the networks are similar to that of the original ARM network (10.76$\pm$0.34 milliseconds). 

To summarise, PrivateNet maintains comparable utility in emotion recognition, while protecting the private attributes from known and unknown adversaries. Also, this approach to privacy preservation has no impact on the inference time of the network.



\begin{table}[t!]
\centering
\small
\setlength\tabcolsep{3pt}
\caption{Inference time of the privacy-preserving network, reported as average~$\pm$ standard deviation (in milliseconds), on a RAF-DB image. These times are similar to those of the original network (10.76~$\pm$ 0.34 milliseconds).}
\vspace{5pt}
\begin{tabular}{lccccccl|l|}
\hline
{\textbf{Protected}} &  \multicolumn{2}{c}{\textbf{Against adversary on}} \\
\cline{2-3}
{\textbf{attribute}} & backbone & ARM \\ \hline
Gender  & 10.86~$\pm$ 0.34& 10.82~$\pm$ 0.35 \\
Age  & 10.77~$\pm$ 0.32 & 10.68~$\pm$ 0.34\\
\hline
\end{tabular}
\label{table: inference_time}
\end{table}

\section{Conclusion}
We addressed the problem of private information  leakage through the  features extracted by  the  intermediate layers of a deep learning classifier. Unlike  works that use an adversarial loss to cause the mis-classification of a private attribute, we obfuscate its associated features using a confusion loss. The proposed approach was validated in a scenario where the goal is to conceal age group and gender attributes from a known adversary with access to the output of the layers of an emotion recognition network. The proposed PrivateNet reduces the accuracy of the adversary to close to that of a random classifier, with negligible effects on the accuracy of the target task.  Moreover, the proposed confusion loss is preferable to the adversarial loss in reducing the leakage of private information with an unknown adversary classifier.

Future work will consider more granular private attributes and the protection of multiple private attributes in a single network. 


\bibliographystyle{IEEEbib}

\end{document}